\documentclass[journal]{IEEEtran}

\usepackage{xcolor,soul,framed} 

\colorlet{shadecolor}{yellow}
\usepackage[pdftex]{graphicx}
\graphicspath{{../pdf/}{../jpeg/}}
\DeclareGraphicsExtensions{.pdf,.jpeg,.png}

\usepackage[cmex10]{amsmath}
\usepackage{array}
\usepackage{mdwmath}
\usepackage{mdwtab}
\usepackage{eqparbox}
\usepackage{url}

\usepackage{amssymb}
\usepackage{multirow}
\newcommand{\RNum}[1]{\uppercase\expandafter{\romannumeral #1\relax}}
\usepackage[ruled]{algorithm2e}
\usepackage{hyperref}
\hypersetup{
    colorlinks=true,            
    linkcolor=blue,             
    filecolor=cyan,             
    urlcolor=magenta,           
    pdftitle={Overleaf Example},
    pdfpagemode=FullScreen,
    }
    
\begin{document}
\bstctlcite{IEEEexample:BSTcontrol}
    \title{Continuous sign language recognition based on cross-resolution knowledge distillation}
  \author{
   \IEEEauthorblockN{Qidan~Zhu$^{1}$, 
    Jing~Li$^{1*}$, 
    Fei~Yuan$^2$, 
    Quan~Gan$^1$}
    
    \IEEEauthorblockA{$^1$ College of Intelligent Systems Science and Engineering, Harbin Engineering University, Harbin, 150001, China}
    \IEEEauthorblockA{$^2$ Northwest Institute of Mechanical and Electrical Engineering, Xianyang, 712099, China}

 \thanks{*Corresponding author\par
Email addresses: \par
zhuqidan@hrbeu.edu.cn (Qidan Zhu), \par
ljing@hrbeu.edu.cn (Jing Li), \par
bohelion@hrbeu.edu.cn (Fei Yuan), \par
gquan@hrbeu.edu.cn (Quan Gan)}
  }

\maketitle

\begin{abstract}
The goal of continuous sign language recognition(CSLR) research is to apply CSLR models as a communication tool in real life, and the real-time requirement of the models is important. In this paper, we address the model real-time problem through cross-resolution knowledge distillation. In our study, we found that keeping the frame-level feature scales consistent between the output of the student network and the teacher network is better than recovering the frame-level feature sizes for feature distillation. Based on this finding, we propose a new frame-level feature extractor that keeps the output frame-level features at the same scale as the output of by the teacher network. We further combined with the TSCM+2D hybrid convolution proposed in our previous study to form a new lightweight end-to-end CSLR network-Low resolution input net(LRINet). It is then used to combine cross-resolution knowledge distillation and traditional knowledge distillation methods to form a CSLR model based on cross-resolution knowledge distillation (CRKD). The CRKD uses high-resolution frames as input to the teacher network for training, locks the weights after training, and then uses low-resolution frames as input to the student network LRINet to perform knowledge distillation on frame-level features and classification features respectively. Experiments on two large-scale continuous sign language datasets have proved the effectiveness of CRKD. Compared with the model with high-resolution data as input, the calculation amount, parameter amount and inference time of the model have been significantly reduced under the same experimental conditions, while ensuring the accuracy of the model, and has achieved very competitive results in comparison with other advanced methods.
\end{abstract}


%
\IEEEpeerreviewmaketitle


\section{Introduction}

\IEEEPARstart{S}{ign} language is a gestural-motor language that can convey semantic information through gestures, hand shapes, facial expressions and body movements to help special people to communicate\cite{rastgoo2021sign}\cite{adaloglou2021comprehensive}\cite{wei2020semantic}. In recent years, some research reports show that the number of hearing-impaired people will increase dramatically in the future, then sign language will occupy an increasingly important position, and CSLR is closer to real scene and has greater practical application value\cite{cui2017recurrent}. Among them, video-based CSLR has attracted extensive attention from researchers because of its advantages such as low cost and no contact\cite{cui2019deep}\cite{huang2021boundary}.\par

Video-based CSLR is the translation of continuous sign language videos into text or speech that can be read and understood by normal people, which can help special people to better integrate into social life\cite{ong2005automatic}\cite{zuo2022c2slr}. Many studies have been carried out by researchers on this subject with good results. In \cite{gao2021rnn}, a Chinese sign language processing method based on RNN converters was proposed by Gao et al. They explored multi-scale visual semantic features by designing a multi-level visual hierarchical transcription network with frame-level, lexical-level and phrase-level BiLstm, and obtained effective recognition results. Recently, Parelli et al.\cite{parelli2022spatio} proposed a sequence learning model based on ST-GCN, which extracts multiple visual representations of sign language actions and captures information about the gesture, shape, appearance and movement of sign language presenters. These models focus more on recognition accuracy, and the complexity and computational effort of the model is often overlooked, or researchers believe that accuracy is the first, which is also correct.\par

\begin{figure*}
  \begin{center}
  \includegraphics[width=3.5in]{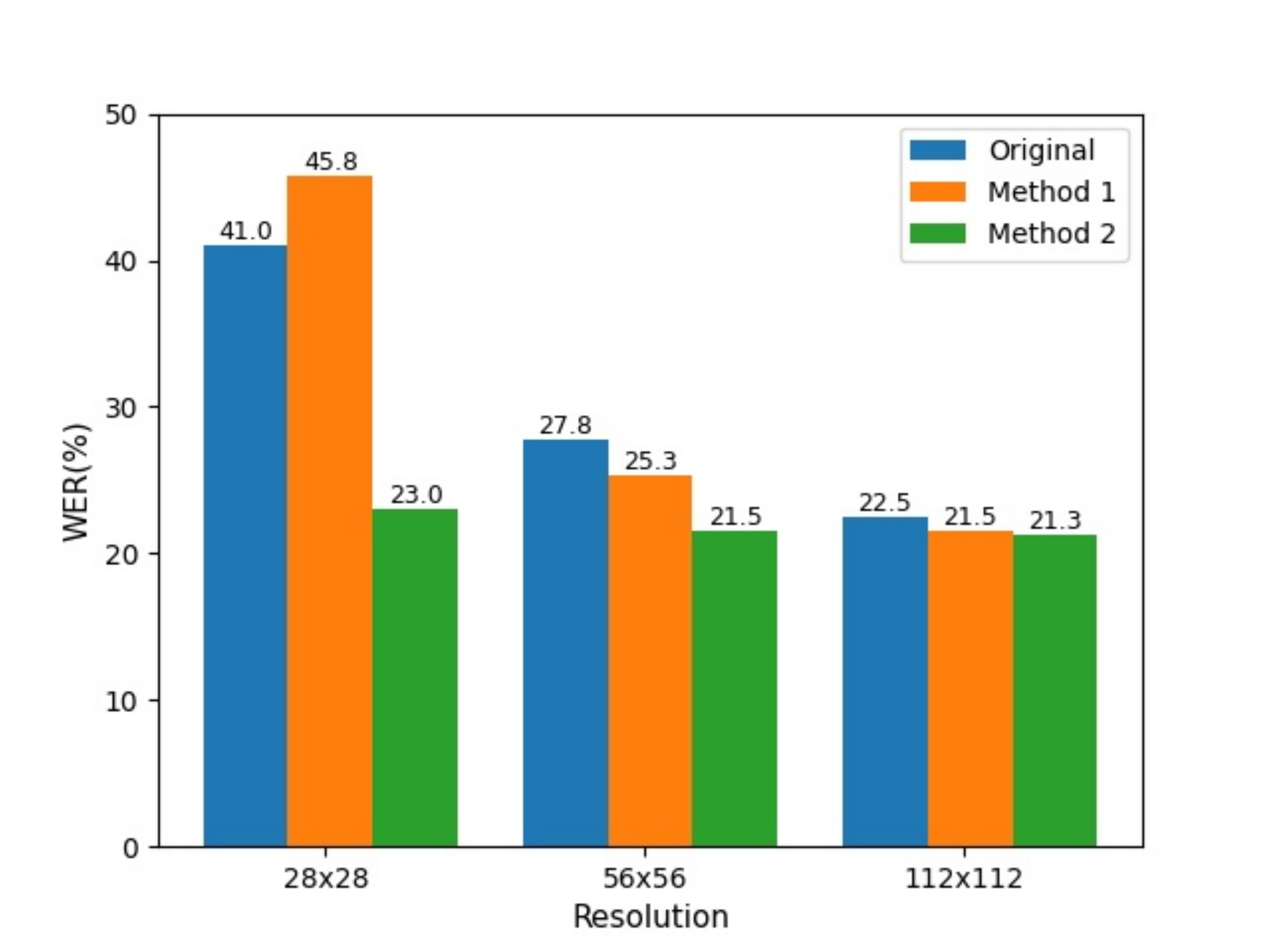}\\
  \caption{Comparison of the experimental results of the original method, method 1 and method 2 on the RWTH dataset. The base model is MSTNet(proposed in our previous study), which is also used as the student model, and the teacher model is MSTNet with 224x224 resolution input. In method 2, only the number of downsampling in the frame-level feature extractor (ResNet) is changed depending on the input resolution, and all models are judged by WER, with the lower the WER the better the recognition.}\label{fig:ljxy1}
  \end{center}
\end{figure*}

However, the ultimate goal of researchers in CSLR should be to apply CSLR models as a communication tool in real-life applications, which can provide considerable practical benefits. Therefore, the real-time requirement of a CSLR model should be considered as the same important position as the accuracy. It is well known that a continuous sign language video represents a combination of multiple sign language words with varying sentence lengths, hence the sheer volume of video data, and the need to fully consider the spatial-temporal relationship between frames when used as a recognition model, leading to a spike in model computation\cite{zhu2023continuous}. Some researchers have moved on to the issue of model real time. We investigated the computational effort of CSLR in \cite{zhu2023continuous} and found that the model computational effort was mainly focused on the pre-frame-level feature extraction. This study has led to a direction for researchers to reduce the computational effort of the model. Recent studies \cite{ma2022rethinking} have shown that low-resolution frames are not necessarily low-quality frames, and experimental validation results demonstrate that low-resolution input is not the main reason for the reduced accuracy of classical recognition model. The mismatch between the input resolution and the input scale of the model is the main reason for the significant reduction of the recognition accuracy. Inspired by \cite{zhu2023continuous}\cite{ma2022rethinking}, we conduct further research with the aim of designing a lightweight CSLR model that can minimize the model computation while minimizing the model accuracy degradation, thus achieving the best compromise between real-time and accuracy of the model. We use cross-resolution knowledge distillation techniques, using low-resolution frames as input to the model, in order to significantly reduce the amount of parameters and calculation of frame-level feature extraction, and obtain the lightweight model.\par

There are two frame-level feature extraction methods used for cross-resolution knowledge extraction. Method 1 is to take a low-resolution frame as input, keep the frame-level feature extractor unchanged, upsample the extracted frame-level features, and then perform feature distillation with the frame-level features output by the teacher network. Method 2 is to reduce the number of downsampling according to the input resolution size, keep the final output frame-level feature size consistent with the frame level feature size output by the teacher network, and followed by feature distillation. In this paper, two methods of cross-resolution knowledge distillation are investigated and compared with the original method (without feature distillation), and the experimental results are shown in Figure 1. It can be seen that the word error rate (WER) of method 2 is 22.8\%, 3.8\% and 0.2\% lower than that of method 1 at the three different resolutions. The lower the resolution of the input, the more obvious is the advantage of method 2. At very low resolutions, the WER results for method 1 are 4.8\% higher than the original method. In summary, for low resolution inputs, method 2 outperforms method 1 in terms of frame-level feature extraction, meaning that Method 2 is more cost-effective in reducing the amount of model computation while maintaining recognition accuracy.\par

Based on the above study, this paper proposes a frame-level feature extraction network for low-resolution input, and designs a new end-to-end CSLR network-LRINet by combining it with the temporal superposition cross module proposed in our previous study \cite{zhu2022temporal}, and then combines LRINet with cross-resolution knowledge distillation and traditional knowledge distillation methods, and introduces stochastic gradient stopping\cite{niu2020stochastic} and the multi-level CTC loss\cite{zhu2022multi} proposed in our previous study for training, constituting our proposed overall model CRKD. CRKD uses high-resolution frames as the input to the teacher network MSTNet(proposed in our previous study) for training, locking the weights after the network is trained, and then using low-resolution frames as input to the student network LRINet to perform knowledge distillation on frame-level features and classification features, respectively. Experiments were conducted on two large-scale continuous sign language datasets and the results demonstrated the effectiveness of CRKD, achieving highly competitive results in comparison with other state-of-the-art methods.\par

\begin{figure*}
  \begin{center}
  \includegraphics[width=5in]{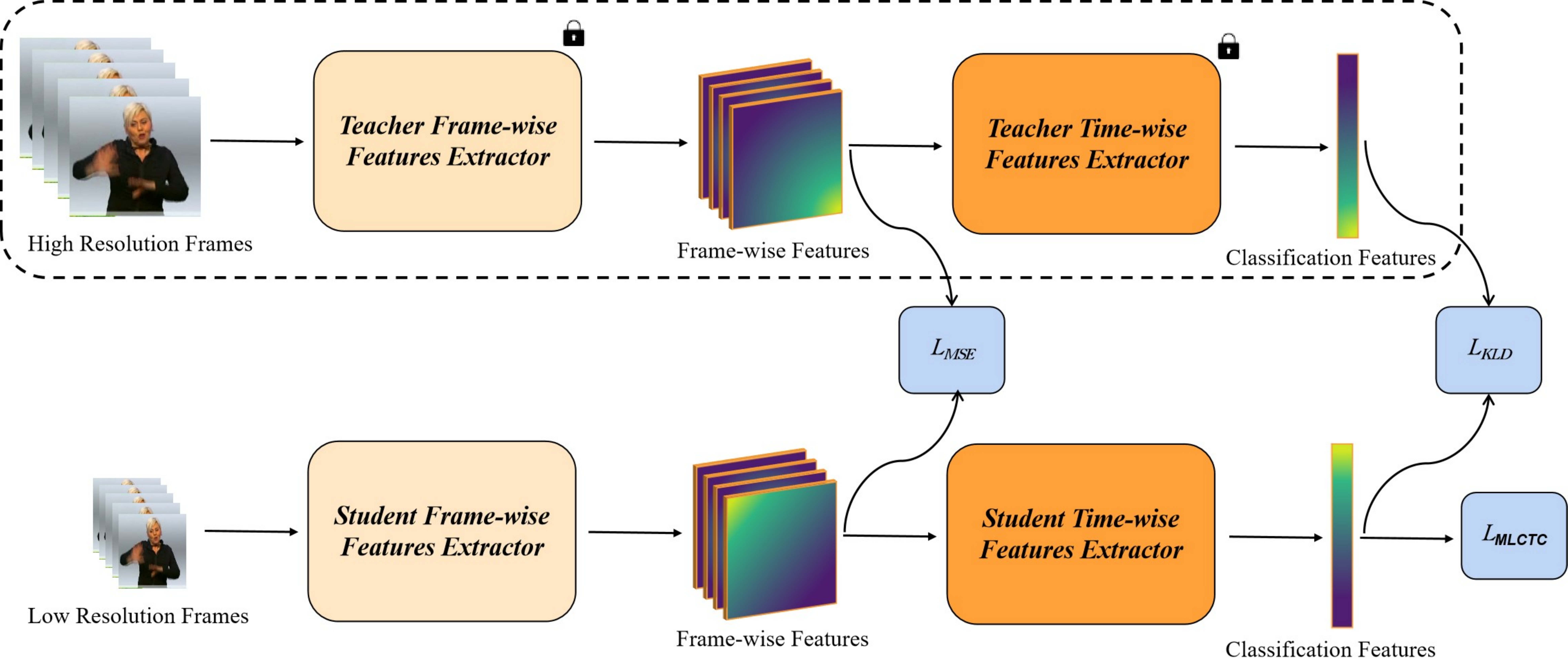}\\
  \caption{Overall architecture of the continuous sign language recognition model based on cross-resolution knowledge distillation.}\label{fig:ljxy2}
  \end{center}
\end{figure*}

The main contributions of this paper are as follows:\par

\begin{itemize}
\item[$\bullet$] In this paper, we find that using cross-resolution knowledge distillation to keep the output frame-level feature scales consistent between the student network and the teacher network for different resolution inputs is better than recovering frame-level feature sizes for feature distillation for feature extraction.
\item[$\bullet$] This paper proposes a new frame-level feature extraction network for low-resolution input video data, which can be combined with the cross-resolution knowledge distillation method to ensure the accuracy of the model while reducing the model computation and the number of parameters.
\item[$\bullet$] This paper proposes a new end-to-end CSLR network LRINet, combining it with cross-resolution knowledge distillation and traditional knowledge distillation methods, and introducing random gradient stopping and multi-stage CTC loss for training, which can able to significantly reduce the computational and parametric quantities of the CSLR model, and ensure the model accuracy.
\item[$\bullet$] The overall CSLR model CRKD proposed in this paper was experimented on two publicly available continuous sign language datasets, and both achieved highly competitive results compared with state-of-the-art methods.
\end{itemize}

\section{Related Work}

\subsection {Continuous Sign Language Recognition}
Video-based CSLR has received a lot of attention from researchers due to its advantages such as non-contact and low cost, with the aim of translating continuous sign language videos into written phrases or spoken words that can be understood by normal people. In CSLR, traditional methods are usually performed by extracting manual features and then combining them with HMM\cite{talukdar2022vision} or dynamic time warping (DTW)\cite{zhang2014threshold} methods. With the development of deep learning, CNN\cite{koller2018deep} has replaced the extraction of manual features, and RNN\cite{al2021deep} or Transform\cite{camgoz2020sign} has replaced HMM and DTW. In the process, researchers have proposed a large number of excellent deep learning models for CSLR\cite{mittal2019modified}\cite{ariesta2018sentence}\cite{de2020sign}\cite{sharma2021asl}. These models focus more on the improvement of model accuracy and better mining of feature information, but the ultimate goal of the field is the need to apply the models in practice, that is, the lightweight of the models.\par

In recent years, several researchers have also focused on improving the real-time performance of recognition models. Cheng et al.\cite{cheng2020fully} proposed a fully convolutional network without pre-training and introduced a jointly trained GFE module to enhance the representativeness of features. Min et al.\cite{min2021visual} proposed visual alignment constraints to predict more alignment supervision by enforcing a feature extractor that makes CSLR networks end-to-end trainable for addressing the overfitting problem of CTC\cite{graves2006connectionist} in sign language recognition. Hu et al.\cite{hu2022transformer} proposed a key frame extraction algorithm for continuous sign language video that adaptively computes the difference threshold and extracts the key frame from the sign language video using image difference and image blur detection techniques. We\cite{zhu2023continuous} proposed a new temporal super-resolution network to reduce the computational effort of the overall CSLR model and improve the real-time performance, and pointed out in the study that the computational effort of the CSLR model is mainly concentrated on the frame-level feature extraction in the early stage. Han et al.\cite{han2022sign} used "2+1D-CNN" for feature extraction and proposed a lightweight temporal channel attention module. We\cite{zhu2022temporal} proposed a zero-parameter, zero-computation temporal superposition crossover module and combined it with 2D convolution to form a hybrid convolution that outperformed other spatial-temporal convolutions in terms of number of parameters, computation and inference time.\par

In order to improve the real-time performance of the model, this paper takes the low-resolution RGB image as the input, uses the lightweight model to extract the frame-level features, and finally uses the hybrid convolution of "TSCM+2D convolution" to establish the space-time relationship.\par

\begin{figure*}
  \begin{center}
  \includegraphics[width=6in]{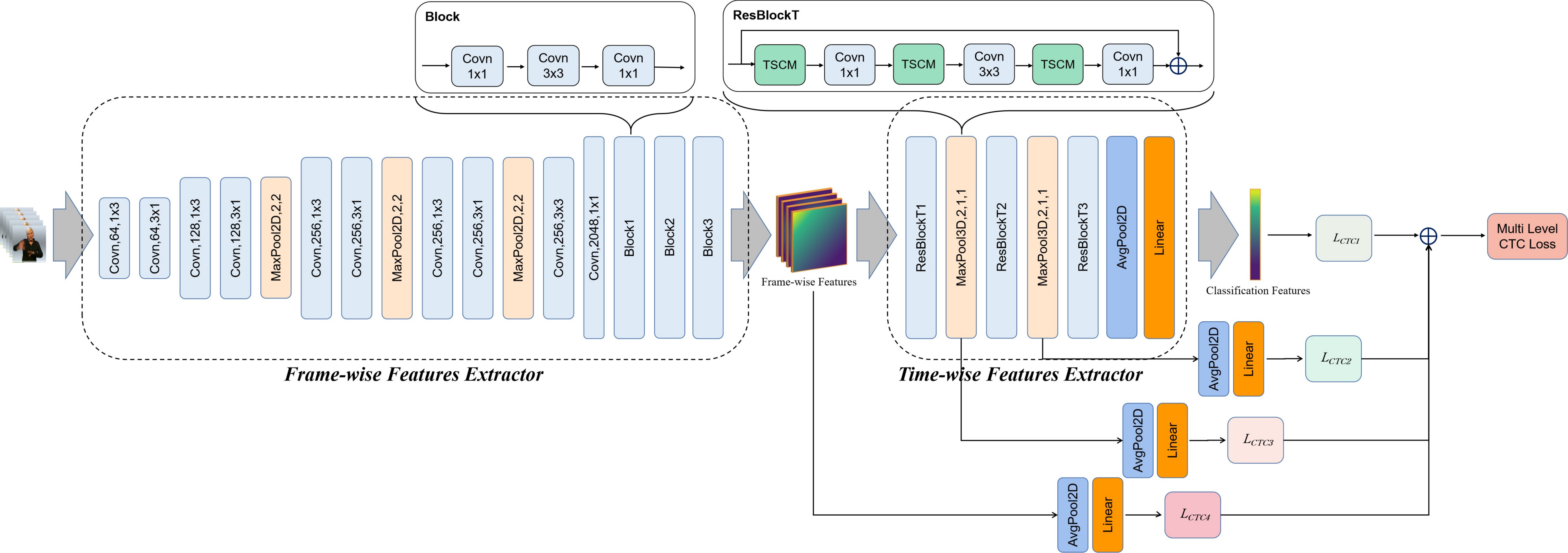}\\
  \caption{Low resolution input network architecture.}\label{fig:ljxy3}
  \end{center}
\end{figure*}

\subsection {Knowledge distillation}
Knowledge distillation is a classical approach for model compression, a concept first introduced by Hinton et al.\cite{hinton2015distilling}, where knowledge transfer is achieved by introducing soft targets related to the teacher's network as part of the final loss in order to induce training of the student's network. After years of development, many branches of knowledge distillation have emerged, which are no longer limited to distillation using the final classification features, including response-based knowledge distillation, feature-based knowledge distillation, and relation-based knowledge distillation. In our approach, we use features with high resolution as input to distill features that use low resolution as input, which falls under feature-based knowledge distillation. Many studies in recent years belong to this type of knowledge distillation method. A 3D-CNN network using knowledge distillation was proposed by Stroud et al. in \cite{stroud2020d3d}, which does not require optical flow branching during inference, but outperforms the dual flow method. The \cite{purwanto2019extreme} is also superior to the dual-stream approach, which is a dual-stream network combined with a new spatiotemporal multi-headed self-attentive mechanism and also uses a super-resolution mechanism to enhance degraded visual information in low-resolution videos, resulting in the construction of an effective framework for extreme low-resolution video action recognition. Ma et al.\cite{ma2022rethinking} took an alternative approach by using cross-resolution knowledge distillation with high-resolution recognition teacher network together with the student network to guide them on the low-resolution framework. In this study, we compress the frame-level feature extractor and the temporal feature extractor of the model based on cross-resolution knowledge distillation and traditional knowledge distillation methods respectively to obtain a lightweight end-to-end model.\par

\section{Methodology}
The overall architecture of CRKD proposed in this paper is shown in Figure 2. The model is mainly composed of two parts: a high-resolution teacher network and a low-resolution student network. In the model training stage, the high-resolution frame is used as the input of the teacher network to train the teacher network, and the weight value is locked after the network is trained. Then, the low-resolution frame is used as the input of the student network, and the knowledge distillation is carried out on the frame-level features and classification features respectively, and finally the student network is trained with multi-level CTC loss. In the testing phase, only low-resolution frames are input into the student network for testing.\par

In the model CRKD, we use the MSTNet\cite{zhu2022multi} proposed in our previous research as the teacher network, and replace the frame-level feature extraction backbone of the network with resnet-50\cite{he2016deep} from resnet-34. This paper uses the proposed low-resolution input network LRINet as the student network, and the details of the proposed network will be described in detail in A.\par

\subsection {Low-resolution input network}

The low-resolution input network LRINet consists of two main components: a frame-level feature extractor and a temporal feature extractor, as shown in Figure 3. The frame-level feature extractor is a 2D-CNN frame-level feature network proposed in this paper for low-resolution inputs. The front part of the network layer reduces the number of parameters and computation by splitting the convolutional kernels, and the back part of the network layer uses a bottleneck structure to keep the number of channels consistent with the teacher network. In this case, the size of the feature map is varied in the first layer of the bottleneck structure by parameter settings of the convolution kernel and padding. The temporal feature extractor uses the "TSCM+2D convolution" hybrid convolution proposed in our previous study [13] to extract temporal features through the residual bottleneck structure.\par

For an input of a continuous sign language video $V=(x_1,x_2,...,x_T)=\{{x_t|_1^T\in \mathbb{R}^{T\times c\times h\times w}}\}$. containing frame $T$, where $x_t$ is the t-th image in the video, $h\times w$ is the size of $x_t$ and $c$ is the number of channels. Let $F_s$  be the frame-level feature extractor and obtain the feature expression as follows:\par

\begin{equation}
f_{spatial} = F_s(V)\in \mathbb{R}^{T\times c^{'}\times h^{'}\times w^{'}}
\end{equation}

Where $f_{spatial}$ is the extracted frame-level feature, $c^{'}$ is the size of the number of channels after frame-level feature extraction, and $h^{'}\times w^{'}$ is the size of the frame-level feature map.\par

The frame-level feature $f_{spatial}$ is then fed into the temporal feature extractor $F_t$:\par

\begin{equation}
f_{temppral} = F_t(f_{spatial})\in \mathbb{R}^{T^{'}\times c^{'}\times h^{'}\times w^{'}}
\end{equation}

Where $f_{temppral}$ is the extracted temporal feature, $T^{'}$ is the down-sampled temporal length of the temporal feature in the time dimension, $c^{'}$ is the size of the number of channels after the temporal feature extraction, and $h^{'}\times w^{'}$ is the size of the temporal feature map. The number of channels and the size of the feature map after the temporal feature extraction are consistent with the previous frame-level features.\par

The final temporal feature $f_{temppral}$ is classified after a global average pooling layer and a fully connected layer:\par

\begin{equation}
f_{calssification} = F_{linear}(F_{avg}(f_{temppral}))\in \mathbb{R}^{T^{'}\times l}
\end{equation}

Where $f_{calssification}$ is the final classification result and $l$ is the number of categories.\par

\subsection {Hybrid loss function}

\textbf{MSE loss.} The mean squared error loss function is a loss function often used in regression problems. In cross-resolution knowledge distillation, we use the mean-squared error loss function for feature distillation to minimize the mean-squared error value between the frame-level features obtained from the teacher model and the student model respectively for the purpose of knowledge transfer, where $n$ is the number of data, $y_i$ is the true value of the i-th data and $y_i^{'}$ is the predicted value of the model output, so minimizing this function is the goal of optimization.\par

\begin{equation}
MSE(y,y^{'}) = \frac{\sum_{\substack{i=1}}^{\substack{n}}(y-y^{'})^{2}}{n} 
\end{equation}

\textbf{KLDiv loss.}\cite{touvron2021training} KL divergence allows comparison of the closeness of two probability distributions and is often used in knowledge distillation. In this paper, we use the model with high resolution inputs as the teacher model and the model with small resolution inputs as the student model, and distill the model by KLDiv loss.\par

\begin{equation}
KLDiv(p,q) = -\sum_{\substack{i=1}}^{\substack{l}}p(x_i)log\frac{q(x_i)}{p(x_i)}
\end{equation}

Where $l$ is the number of data, $x_i$ is the i-th data, $p(x_i)$ is the classification probability output by the student model and $q(x_i)$ is the classification probability output by the teacher model.\par

\textbf{Multi-level CTC loss.}\cite{zhu2022multi} CTC introduces a blank label {-} to mark unclassified labels during decoding. For the T frames of the input video $V$, the label for each frame is represented by $\pi=(\pi_1,\pi_2,...,\pi_T)$, where $\pi_t\in v\cup \{-\}$, $v$ are sign language vocabulary, the posterior probability of the label is:\par

\begin{equation}
p(\pi|V)=\prod_{\substack{t=1}}^{\substack{T}}p(\pi_t|V)=\prod_{\substack{t=1}}^{\substack{T}}Y_{t,\pi_t}
\end{equation}

For a given sentence-level label $s=(s_1,s_2,...,s_L)$, where $L$ is the number of words in the sentence. Then the conditional probability of label $s$ is the sum of the probabilities of occurrence of all corresponding paths.\par

\begin{equation}
p(s|V)=\sum_{\substack{\pi\in B^{-1}(s)}}p(\pi|V)
\end{equation}

Where $B^{-1}(s)=\{\pi|B(\pi)=s\}$ is the inverse mapping of $B$ and $B$ is a many-to-one mapping defined by CTC. The CTC loss is defined as the negative log likelihood of the conditional probability of the label $s$:\par

\begin{equation}
L_{CTC}=-\ln p(s|V)
\end{equation}

The multi-level CTC loss can then be expressed as follows, where $m$ is the number of CTCs.\par

\begin{equation}
\begin{aligned}
    L_{sum}&=-\ln \prod_{\substack{i=1}}^{\substack{m}} p(s|V_i)\\
    &=-\ln (p(s|V_1)p(s|V_2)...p(s|V_{m-1})p(s|V_m))
\end{aligned}
\end{equation}

\textbf{Hybrid loss.} In this paper, we use a mixture of losses for training, with different losses having their different functions. MSE loss minimizes the error between the frame-level features of the teacher model and the frame-level features of the student model when frame level features are distilled. KLDiv loss uses soft labels to train the student model, transferring knowledge from the larger model to the smaller model. Multi-level CTC loss is used to train the model using real labels. The different levels of CTC loss not only allow for better decoding of temporal features, but also allow for a good update of the parameters of the shallow network. Hybrid loss combine their advantages.\par

\begin{equation}
L_{Mix}= \sum_{\substack{i=1}}^{\substack{m}}L_{CTC_i}+\alpha L_{MSE}+\beta L_{KLDiv}
\end{equation}

Where $\alpha$, $\beta$ are two hyperparameters, and the weight of MES loss and KLDiv loss in the hybrid loss is changed by changing the values of these two hyperparameters.\par

\subsection {Model parameters}

The low resolution input network is proposed to accommodate video with low-resolution frames as input. It keeps the frame-level feature output size consistent with the teacher model with high resolution as input, which is convenient for feature distillation. We take the video with a length of 200 frames and a resolution of $3\times 72\times 72$ as the input. The overall architecture of the low-resolution input network is shown in Table \RNum{1}, where   is the number of channels in the frame-level feature output of the teacher network. This low-resolution input network consists of two main components: a frame-level feature extractor and a temporal feature extractor.\par

In the frame-level feature extractor, the front part of the network layer uses split convolutional kernels to reduce the number of parameters and computation, and the back part of the network layer uses a bottleneck structure to match the number of frame-level feature channels in the output of the teacher model. In the first bottleneck structure, the 2D convolution is used to match the output feature size of the teacher model by varying the size of the convolution kernel and the padding parameter according to the input resolution. The input resolution is based on $56\times 56$, the convolution kernel is based on $1\times 1$, the padding is always 0, and the resolution and convolution kernel size are increased sequentially in steps of 16 and 2 respectively. For example, when the input resolution is $56\times 56$, the first 2D convolution kernel size is 1 and padding is 0. When the input resolution is $72\times 72$, the first 2D convolution kernel size is 3 and padding is 0, and so on. The relationship between the convolution kernel parameter setting and the resolution is shown in equation $(11)$, where $k$ is the convolution kernel size, $r$ is the input resolution, and the value 8 indicates that three spatial down-sampling in steps of 2 have been performed, and the relationship between the four resolutions and the convolution kernel is shown in Table \RNum{2}.\par

\begin{equation}
k= \frac{r}{8}-6,r\in [56,72,88,104...]
\end{equation}

In the temporal feature extractor, we use the TSCM+2D hybrid convolution as the base unit, and the residual bottleneck structure as the base block, with the number of channels aligned with the number of channels in the final output of the frame-level features, using a total of 3 layers of residual bottleneck structure. In this case, we perform two maximum pooling operations, only pooling in the time dimension, and use batch normalization and ReLu activation functions after each convolution operation throughout the network.\par

\begin{table*}[!htbp]
\centering
\caption{Main architecture of the low-resolution input network, where $N$ is the number of frame-level feature channels at the output of the teacher's network}
\label{tab:aStrangeTable1}
\begin{tabular}{c|c|c|c|c|c|c}
\hline  
Layer& Input Size& Input Channels& Middle Channels& Output Channels& Kernel Size& Stride\\
\hline  
Conv& $200\times 3\times 72\times 72$& 3& -& 64& $1\times 3$& 1\\
\hline  
Conv& $200\times 64\times 72\times 72$& 64& -& 64& $3\times 1$& 1\\
\hline  
Conv& $200\times 64\times 72\times 72$& 64& -& 128& $1\times 3$& 1\\
\hline  
Conv& $200\times 128\times 72\times 72$& 128& -& 128& $3\times 1$& 1\\
\hline  
Max2D& $200\times 128\times 72\times 72$& 128& -& 128& $2\times 2$& 2\\
\hline  
Conv& $200\times 128\times 36\times 36$& 128& -& 256& $1\times 3$& 1\\
\hline  
Conv& $200\times 256\times 36\times 36$& 256& -& 256& $3\times 1$& 1\\
\hline  
Max2D& $200\times 256\times 36\times 36$& 256& -& 256& $2\times 2$& 2\\
\hline  
Conv& $200\times 256\times 18\times 18$& 256& -& 256& $1\times 3$& 1\\
\hline  
Conv& $200\times 256\times 18\times 18$& 256& -& 256& $3\times 1$& 1\\
\hline  
Max2D& $200\times 256\times 18\times 18$& 256& -& 256& $2\times 2$& 2\\
\hline  
Conv& $200\times 256\times 9\times 9$& 256& -& 256& $3\times 3$& 1\\
\hline  
Conv& $200\times 256\times 9\times 9$& 256& -& $N$& $1\times 1$& 1\\
\hline  
Bottleneck1& $200\times N\times 9\times 9$& $N$& 128& $N$& -& 1\\
\hline  
Bottleneck2& $200\times N\times 7\times 7$& $N$& 128& $N$& -& 1\\
\hline  
Bottleneck3& $200\times N\times 7\times 7$& $N$& 256& $N$& -& 1\\
\hline  
ResBottleneck1& $200\times N\times 7\times 7$& $N$& 128& $N$& -& 1\\
\hline  
Max3D& $200\times N\times 7\times 7$& $N$& -& $N$& $2\times 1\times 1$& $2\times 1\times 1$\\
\hline  
ResBottleneck2& $100\times N\times 7\times 7$& $N$& 128& $N$& -& 1\\
\hline  
Max3D& $200\times N\times 7\times 7$& $N$& -& $N$& $2\times 1\times 1$& $2\times 1\times 1$\\
\hline  
ResBottleneck3& $50\times N\times 7\times 7$& $N$& 256& $N$& -& 1\\
\hline  
GlobalAvgPool& $50\times N\times 7\times 7$& $N$& -& $N$& -& -\\
\hline  
FC& $50\times N\times 1\times 1$& $N$& -& 1296& -& -\\
\hline  
Softmax& $50\times N\times 1\times 1$& 1296& -& 1296& -& -\\
\hline  
\end{tabular}
\end{table*}

\begin{table*}[!htbp]
\centering
\caption{Relationship between 2D convolution parameter setting and input resolution in the first bottleneck}
\label{tab:aStrangeTable2}
\begin{tabular}{c|c|c|c}
\hline  
Input Size& Kernel Size& Stride& Padding\\
\hline  
$56\times 56$& $1\times 1$& 1& 0\\
\hline  
$72\times 72$& $3\times 3$& 1& 0\\
\hline  
$88\times 88$& $5\times 5$& 1& 0\\
\hline  
$104\times 104$& $7\times 7$& 1& 0\\
\hline  
\end{tabular}
\end{table*}

\section{Experiment}

In this section, we conduct comprehensive experiments on two widely used CSLR datasets to validate the effectiveness of the proposed model in this paper. A series of ablation experiments are also conducted to demonstrate the role of each component of the proposed model.\par

\subsection {Dataset}

RWTH-PHOENIX-Weather-2014 (RWTH) dataset\cite{koller2015continuous}: RWTH is recorded by a public weather radio and television station in Germany. All the presenters are dressed in dark clothing and perform sign language in front of a clean background. The video in this dataset is recorded by 9 different presenters, with a total of 6841 different sign language sentences (including 77321 sign language word instances and 1232 words). All videos are pre-processed to a resolution of $210\times 260$ and a frame rate of 25 frames per second (FP/S). The dataset is officially divided into 5672 training samples, 540 validation samples and 629 test samples.\par

Chinese Sign Language (CSL) dataset\cite{huang2018video}: CSL contains 100 sentences of Chinese everyday language, each sentence is demonstrated 5 times by 50 presenters, and the vocabulary size is 178. The video resolution is $1280\times 720$ and the frame rate 30 FP/S. The dataset includes two ways to divide the training and test sets: split I and split II. This paper is an experimental study on split I.\par

\begin{table*}[!htbp]
\centering
\caption{We compare performance with different CSLR models on the RWTH dataset, using WER as the metric (lower is better), where Full indicates recognition using only the full RGB image and Extra clues indicates that other cues are used for recognition}
\label{tab:aStrangeTable3}
\begin{tabular}{c|c|c|c|c|c}
\hline  
\multirow{2}{*}{Methods}& \multirow{2}{*}{Backbone}& \multirow{2}{*}{Full}& \multirow{2}{*}{Extra clues}& \multicolumn{2}{c}{WER(\%)}\\
\cline{5-6}
 & & & & Dev& Test\\
\hline
Re-Sign\cite{koller2017re}& GoogLeNet& Y& -& 27.1& 26.8\\
\hline  
CNN+LSTM+HMM\cite{koller2019weakly}& GoogLeNet& -& Y& 26.0& 26.0\\
\hline  
FCN\cite{cheng2020fully}& Custom& Y& -& 23.7& 23.9\\
\hline  
CMA\cite{pu2020boosting}& GoogLeNet& -& Y& 21.3& 21.9\\
\hline  
VAC\cite{min2021visual}& ResNet18& Y& -& 21.2& 22.3\\
\hline  
ResNetT34\cite{zhu2022temporal}& ResNet34& Y& -& 21.1& 21.1\\
\hline  
STMC\cite{zhou2020spatial}& VGG11& -& Y& 21.1& 20.7\\
\hline  
MSTNet\cite{zhu2022multi}& ResNet34& Y& -& 20.3& 21.4\\
\hline  
HST-GNN\cite{kan2022sign}& ResNet152& -& Y& 19.5& 19.8\\
\hline  
H-GAN\cite{elakkiya2021optimized}& Custom& -& Y& 18.8& 20.7\\
\hline  
CRKD& Custom& Y& -& 21.4& 20.9\\
\hline  
\end{tabular}
\end{table*}

\begin{table*}[!htbp]
\centering
\caption{We use spilt I to divide the CSL dataset and conduct experiments to compare performance with different CSLR models, using WER as a metric (lower is better), where Full indicates that only the full RGB image is used }
\label{tab:aStrangeTable4}
\begin{tabular}{c|c|c|c}
\hline  
Methods& Full& Extra clues& WER(\%)\\
\hline
SubUNet\cite{cihan2017subunets}& Y& -& 11.0\\
\hline
Align-iOpt\cite{pu2019iterative}& Y& -& 6.1\\
\hline
DPD\cite{zhou2019dynamic}& Y& -& 4.7\\
\hline
SF-Net\cite{yang2019sf}& Y& -& 3.8\\
\hline
FCN\cite{cheng2020fully}& Y& -& 3.0\\
\hline
CrossModal\cite{papastratis2020continuous}& -& Y& 2.4\\
\hline
STMC\cite{zhou2020spatial}& -& Y& 2.1\\
\hline
SLRGAN\cite{papastratis2021continuous}& -& Y& 2.1\\
\hline
MSTNet\cite{zhu2022multi}& Y& -& 0.7\\
\hline
CRKD& Y& -& 2.6\\
\hline  
\end{tabular}
\end{table*}

\subsection {Implementation Rules}

In the implementation of the overall model experiment in this paper, the training uses the Adam optimizer\cite{kingma2014adam}, the initial learning rate and weight factor are set to $10^{-4}$, and the batch size used is 2. We use MSTNet as the teacher network and ResNet-50 for the frame-level feature extraction backbone. The graphics card used in this experiment is RTX3090Ti, and the size of GPU dedicated memory is 24G.\par

When training on the RWTH dataset, data augmentation is performed using random cropping and random flipping. For random cropping, the input data size is $256\times 256$, and the size after random cropping is $224\times 224$. For random flipping, the flipping probability is set to 0.5. The flipping and cropping processes are performed for the video sequences. In addition, a temporal enhancement process is performed to randomly increase or decrease the length of the video sequence within $\pm 20\%$. After data augmentation, the input data is uniformly scaled to a specified low resolution. The model is trained using 4-level CTC loss. In the hybrid loss, the values of $\alpha$ and $\beta$ are 140 and 120 respectively. In the model test, only the center clipping is used for data augmentation, and the beam search algorithm is used for decoding in the final CTC decoding stage, with the beam width is 10. In the model test, only the center clipping is used for data augmentation. In the final CTC decoding stage, the beam search algorithm is used for decoding, and the beam width is 10. There are 85 epochs in the training stage, and the learning rate decreases by 80\% at the 45th and 65th epochs.\par

When training on the CSL dataset, only random cropping is used for data augmentation and the model is trained using 2-level CTC loss. For testing, only central cropping is used for data augmentation. There are 30 epochs in the training phase, and the learning rate decreases by 90\% in the 20th epoch, and the values of $\alpha$ and $\beta$ in the hybrid loss are 280 and 140 respectively.\par

\subsection {Judgment Criteria}

WER\cite{koller2015continuous} is widely used as a criterion in CSLR. It is the sum of the minimum number of insertion, substitution, and deletion operations required to convert a recognized sequence into a standard reference sequence. Lower WER means better recognition performance, and its definition is as follows:\par

\begin{equation}
WER=100\%\times \frac{ins+del+sub}{sum}
\end{equation}

Where $ins$ represents the number of words to be inserted, $del$ represents the number of words to be deleted, $sub$ represents the number of words to be replaced, and $sum$  represents the total number of words in the label.\par

\subsection {Experimental Results}

The proposed model CRKD is experimented on two publicly available datasets, the RWTH dataset and the CSL dataset. The model recognition accuracy is shown in Table \RNum{3} and Table \RNum{4}, and the curves generated from the WER in Table \RNum{3} and Table \RNum{4} are shown in Figure 4 and Figure 5.\par

As can be seen in Table \RNum{3}, CRKD achieves highly competitive results on the RWTH dataset compared to other state-of-the-art models, with WERs of 21.4\% and 20.9\% on the validation and test sets, respectively. It can be seen from Table \RNum{2} that CRKD also achieves competitive results on the CSL dataset, with a WER of 2.6\% on the test set. As can be seen in Figure 4 and Figure 5, WER decreases with the increase of epoch, and WER decreases significantly when the first lr changes. As can be seen in Figure 4 and Figure 5, WER decreases with the increase of epoch, and WER decreases significantly when the first lr changes. On the RWTH dataset, WER reaches the minimum value of 21.4\% at the 68th epoch for the validation set, and 20.9\% at the 71st epoch for the test set; On the CSL dataset, WER reaches the minimum value of 2.6\% at the 27th epoch of the test set.\par

\begin{figure*}
\begin{minipage}{0.5\textwidth}
\includegraphics[width=3.5in,height=2.45in]{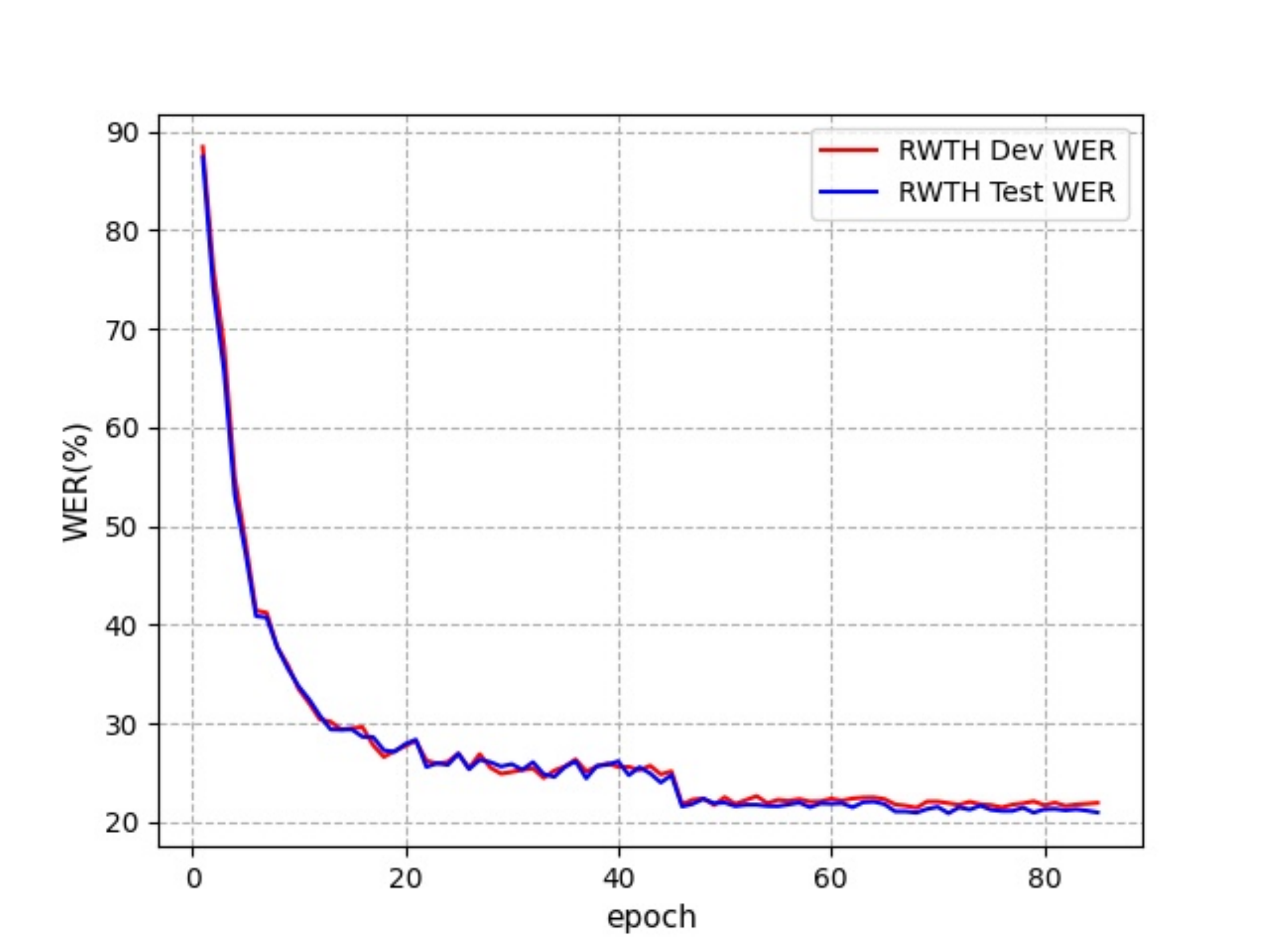}
\caption{WER change curves for the validation and test sets of the RWTH dataset.}
\label{fig:ljxy4}
\end{minipage}
\hfill
\begin{minipage}{0.5\textwidth}
\includegraphics[width=3.5in,height=2.33in]{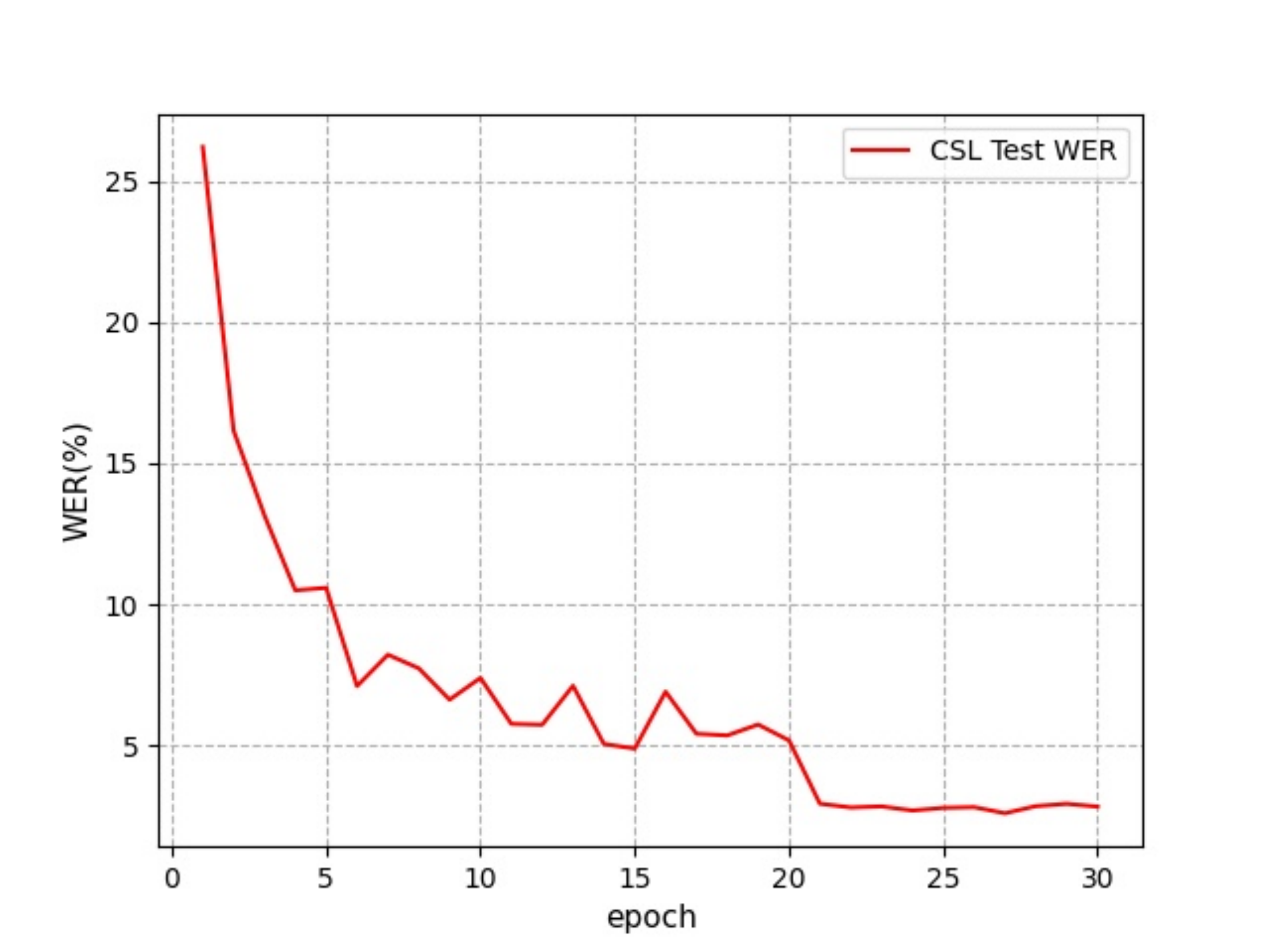}
\caption{WER change curves for the test set of the CSL dataset.}
\label{fig:ljxy5}
\end{minipage}
\end{figure*}

\begin{table*}[!htbp]
\centering
\caption{Comparison of different models in terms of number of parameters, computational effort, inference time and accuracy}
\label{tab:aStrangeTable5}
\begin{tabular}{c|c|c|c|c|c}
\hline  
Model type& CRKD& ResNetT34\cite{zhu2022temporal}& MSTNet\cite{zhu2022multi}& VAC\cite{min2021visual}& CNN+BiLSTM+CTC\\
\hline  
BackBone& Custom& ResNet34& ResNet34& ResNet18& ResNet18\\
\hline  
parameter(M)& 10.5& 22.0& 120.3& 33.0& 15.0\\
\hline  
parameter memory(MB)& 40.1& 83.9& 458.8& 125.9& 57.2\\
\hline  
calculation quantity(GFlops)& 257.0& 671.1& 736.5& 366.0& 365.1\\
\hline  
inference time(ms)& 46.9& 81.9& 75.0& 43.7& 45.9\\
\hline  
WER(\%)& 20.9& 21.1& 21.4& 22.3& 26.7\\
\hline  
\end{tabular}
\end{table*}

\begin{figure*}
\begin{minipage}{0.5\textwidth}
\includegraphics[width=3.5in,height=2.45in]{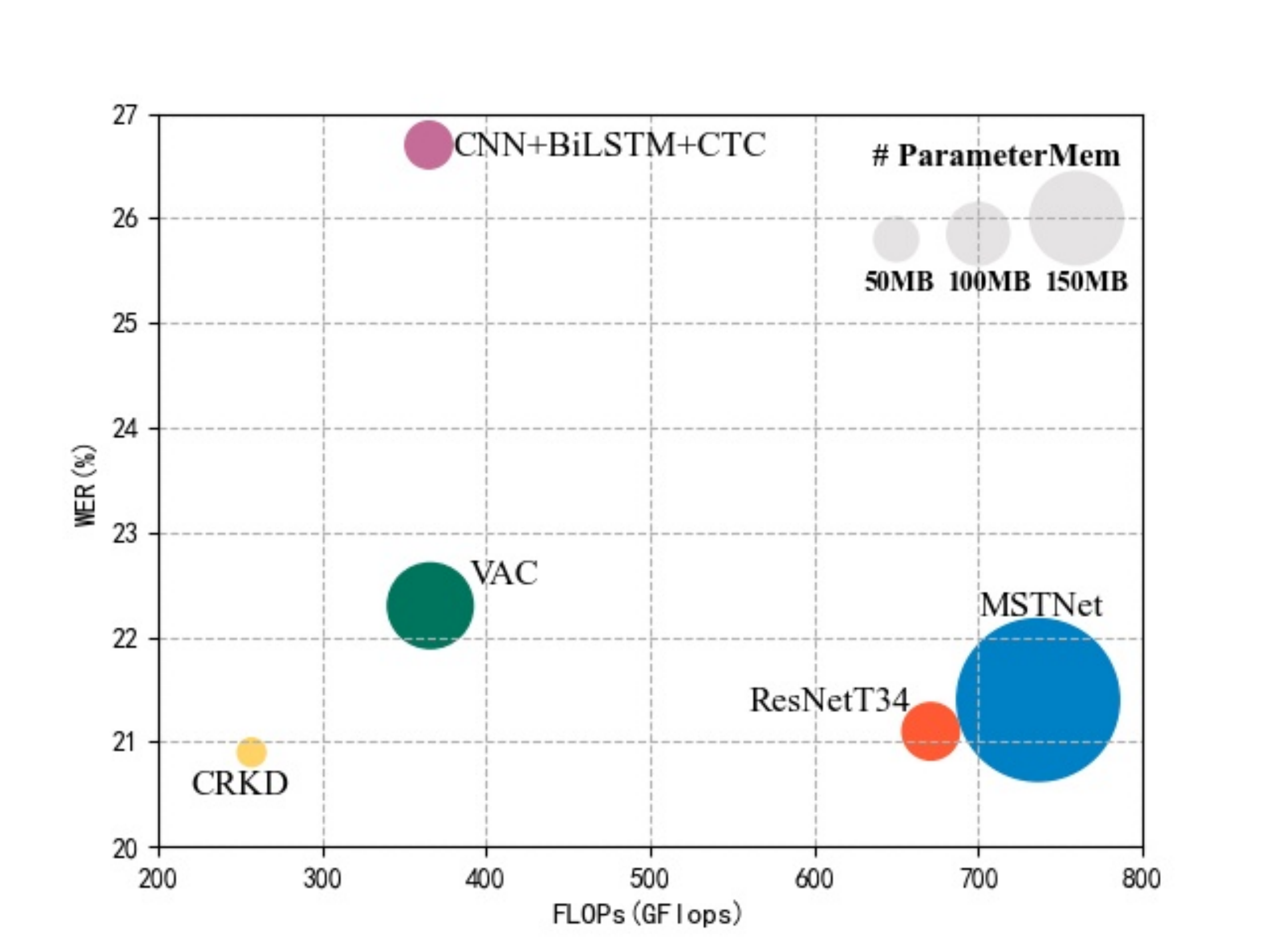}
\caption{Scatter plots of different models in terms of parameter memory, computational effort and accuracy.}
\label{fig:ljxy6}
\end{minipage}
\hfill
\begin{minipage}{0.5\textwidth}
\includegraphics[width=3.5in,height=2.33in]{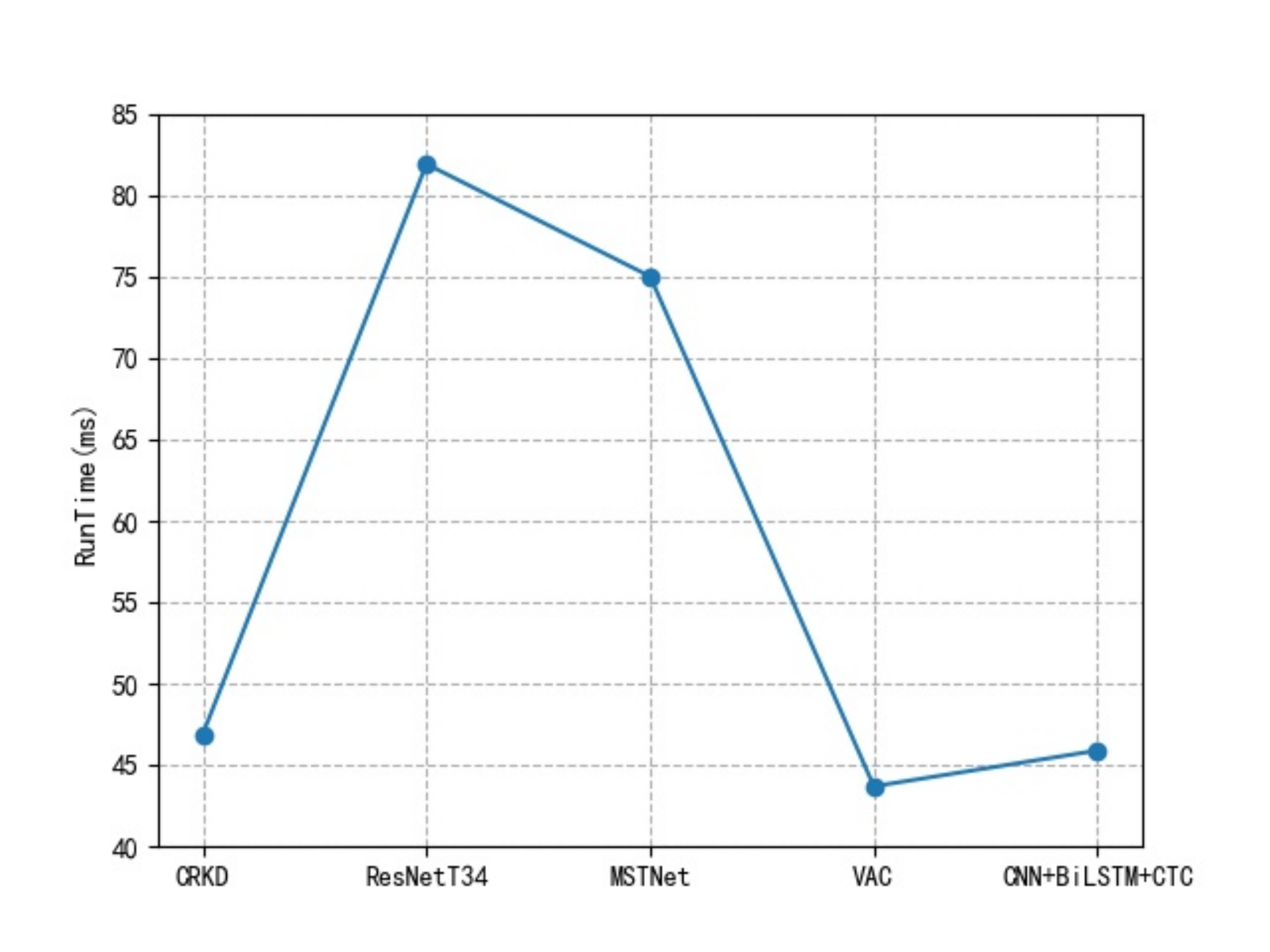}
\caption{Plots of different models in terms of inference time.}
\label{fig:ljxy7}
\end{minipage}
\end{figure*}

\subsection {Real-time performance of the model}

In order to further analyze the real-time performance of the proposed model CRKD in this paper, we experimentally compare the model in this paper with five other models in terms of number of parameters, computation, inference time and accuracy under the same experimental setup to demonstrate the superiority of CRKD in terms of real-time performance.\par

The computational effort of each model is strongly correlated with the size and quantity of the input data. In our model CRKD, the input resolution of LRINet is set to $3\times 72\times 72$ and the video frame temporal length is 200 frames; the input data size of other models is $3\times 224\times 224$ and the video frame temporal length is also 200 frames, and the number of parameters and computation of different models are calculated under the above conditions for experimental comparison. For each model, we perform experimental validation on the RWTH dataset and count the mean inference time as a reference, with WER as the accuracy indicator, where a smaller WER means higher accuracy. The inference time statistics are not only related to the model itself but also closely related to the device used. The graphics card used for the experiments in this paper is RTX3090Ti, and the experimental results are shown in Table \RNum{5}, while the data in Table \RNum{5} are visualized in Figures 6 and 7.\par

\begin{table*}[!htbp]
\centering
\caption{Ablation study with different temporal extractors}
\label{tab:aStrangeTable6}
\begin{tabular}{c|c|c}
\hline  
model& transformer& hybrid convolution\\
\hline  
dev& 20.9& 21.4\\
\hline  
test& 21.5& 20.9\\
\hline  
\end{tabular}
\end{table*}

\begin{table*}[!htbp]
\centering
\caption{Ablation study with different loss functions}
\label{tab:aStrangeTable7}
\begin{tabular}{c|c|c|c|c}
\hline  
loss function& CTC& multi-level CTC& multi-level CTC+MSE& multi-level CTC+MSE+KLD\\
\hline  
dev& 25.2& 22.5& 21.6& 21.4\\
\hline  
test& 25.1& 22.7& 21.3& 20.9\\
\hline  
\end{tabular}
\end{table*}

\begin{table*}[!htbp]
\centering
\caption{Ablation study with different input resolutions}
\label{tab:aStrangeTable8}
\begin{tabular}{c|c|c|c|c}
\hline  
resolution& $56\times 56$& $72\times 72$& $88\times 88$& $104\times 104$\\
\hline  
parameter(M)& 10.4& 10.5& 10.8& 11.2\\
\hline  
calculation quantity(GFlops)& 170.4& 257.0& 366.3& 498.1\\
\hline  
inference time(ms)& 32.8& 46.9& 63.9& 84.0\\
\hline  
dev& 21.8& 21.4& 20.9& 21.3\\
\hline  
test& 22.0& 20.9& 21.1& 21.1\\
\hline  
\end{tabular}
\end{table*}

It can be seen from Table \RNum{5} that: 1) CRKD is the smallest compared to other models in terms of the number of parameters and computation, but the inference time is longer than VAC, because there is a shift operation in the "TSCM+2D" hybrid convolution used, which increases the inference time of the model; 2) the inference time of CRKD is very similar to that of CNN+ BiLSTM+CTC, but the number of parameters in CRKD is 29.9\% lower than that in CNN+BiLSTM+CTC, and the WER is reduced by 5.8\%; 3) CRKD is only 0.2\% lower in WER than ResNetT34, but the number of parameters is reduced by 52.2\%, the computation is reduced by 61.7\%, and the inference time is reduced by 42.7\%.\par

It can be seen from Figure 6 that the CRKD proposed in this paper, relative to other models, is optimal in terms of model accuracy, number of parameters and computational effort. It can also be seen from Figure 7 that although the inference time of CRKD is not the smallest, it only differs from the smallest VAC model by 3.2ms. Therefore, the proposed CRKD achieves a balance of performance and accuracy with a smaller number of parameters, which can make the deployment cost of the model lower, reduce the inference time and ensure the real-time performance of the model.\par

\subsection {Ablation Experiment}
In this section, we conducted ablation experiments on the RWTH dataset to further validate the effectiveness of each component of the model CRKD. In the ablation experiment, the input resolution is set to $3\times 72\times 72$, and WER is used as the metric. The smaller WER is, the better the performance.\par

\textbf{Ablation of different temporal feature extractors.} We use two different temporal feature extractors combined with the same frame-level feature extractor to form different models and compare their accuracy. The ablation effects of different temporal feature extractors are shown in Table \RNum{6}. The two temporal feature extractors are the MSTNet temporal feature extraction part and the temporal feature extraction part composed of "TSCM+2D convolution" hybrid convolution. As can be seen from Table \RNum{6}, in terms of the model accuracy of the test set, the second temporal feature extractor is 0.6\% higher than the first temporal feature extractor. That is, under the current parameter setting conditions, the temporal and spatial modeling ability of the temporal feature extractor composed of hybrid convolution is better than that of the temporal feature extraction part of MSTNet.\par 

\textbf{Ablation of different loss functions.} The ablation effect when using different loss functions is shown in Table \RNum{7}. Different loss functions have different functions. Multi-level CTC loss is to use real labels to train the model. Different levels of CTC loss can not only better decode the temporal features, but also make the parameters of the shallow network well updated. When MSE loss distills the frame-level features, the error between the frame-level features of the teacher model and the frame-level features of the student model is minimized. KLDiv loss uses soft labels to train the student model and transfer the knowledge of the large model to the small model. It can be seen from Table \RNum{7} that the model recognition accuracy is the worst when CTC loss is used only, at which point the WERs for the validation and test sets are 25.2\% and 25.1\% respectively. With the increase of the loss function, the WER of the network decreases. Finally, when using the hybrid loss function, the WER of the model is 3.8\% lower than that of the original CTC loss in the validation set and 4.2\% lower than that of the test set.\par

\textbf{Ablation of different input resolutions.} The ablation effect when the input data are different input resolutions is shown in Table \RNum{8}. The ablation experiment includes four levels: parameter quantity, calculation quantity, inference time and accuracy. It is calculated using different resolutions as the input of LRINet with a video frame length of 200 frames. In the ablation experiment, it is found that with the increase of resolution, the parameters of the model changed little, while the calculation and inference time increased linearly. At a resolution of $72\times 72$, the model accuracy reaches a minimum of 21.4\% and 20.9\% for the validation and test set respectively, while the number of parameters and inference time of the model are 10.5M and 46.9ms respectively. The amount of model parameters at this resolution is basically the same as that at resolution $56\times 56$. Although the inference time is 14.1ms longer than that at resolution $56\times 56$, WER is 1.1\% lower.

\section{Conclusion}

In the research on the real-time performance of the CSLR, this paper found that using the cross-resolution knowledge distillation method to keep the frame-level feature scale of the output of the student network and the teacher network consistent under the condition of different resolution input is better than recovering the frame-level feature size for feature distillation. In view of the above findings, this paper proposes a new frame-level feature extraction network suitable for low-resolution input, and further combines it with "TSCM+2D convolution" hybrid convolution to propose a new end-to-end CSLR network LRINet. We use LRINet combine cross-resolution knowledge distillation and traditional knowledge distillation to form a CSLR model CRKD. The experiments show that CRKD has advantages over other models in terms of parameters, computation and inference time while ensuring accuracy, and achieves the balance between performance and accuracy. The reduction of parameter quantity and calculation quantity reduces the deployment cost of the model, and the reduction of inference time ensures the real-time of the model. \par

Our model CRKD is proposed for low-resolution input, which can greatly reduce the number of parameters and computational effort, but does not take into account the redundancy between video frames. Using sparse video data as input can significantly reduce the amount of computation and improve the inference time. Therefore, it is worth studying to use fewer video frames to obtain the same or better sign language recognition effect as the current available. \par

\section*{Acknowledgment}

This work was supported in part by the Development Project of Ship Situational Intelligent Awareness System, China under Grant MC-201920-X01, in part by the National Natural Science Foundation of China under Grant 61673129. \par

~\\\par
\textbf{Data availability} The datasets used in the paper are cited properly.\par

\section*{Declarations}

\textbf{Conflict of interest} The authors declare that they have no known competing financial interests or personal relationships that could have appeared to influence the work reported in this paper.\par


%





\ifCLASSOPTIONcaptionsoff
  \newpage
\fi





\bibliographystyle{IEEEtran}
\bibliography{IEEEabrv,Bibliography}

\vfill


\end{document}